\documentclass{article}

    \PassOptionsToPackage{numbers, compress}{natbib}

\usepackage[main, final]{neurips_2026}

\usepackage[utf8]{inputenc} 
\usepackage[T1]{fontenc}    
\usepackage{url}            
\usepackage{booktabs}       
\usepackage{amsfonts}       
\usepackage{nicefrac}       
\usepackage{microtype}      
\usepackage{amsmath}
\usepackage{amsthm}
\usepackage{amssymb}
\usepackage{multirow}
\usepackage{graphicx}
\usepackage{tabularx}
\usepackage{wrapfig}  
\usepackage{lipsum}
\usepackage{subcaption}
\usepackage[table,xcdraw]{xcolor}
\usepackage{mathtools}
\usepackage{enumitem}
\usepackage{tikz}
\usetikzlibrary{tikzmark, calc}
\usepackage[ruled,vlined]{algorithm2e}
\usepackage{bbm}
\usepackage{dsfont}
\newtheorem{definition}{Definition}
\newtheorem{proposition}{Proposition}
\newcolumntype{C}{>{\centering\arraybackslash}X}

\usepackage[colorlinks, linkcolor=mydarkred, citecolor=gray]{hyperref}
\definecolor{mydarkred}{rgb}{0.6,0,0}
\definecolor{myblue}{HTML}{268BD2}
\definecolor{mygreen}{HTML}{658354}

\usepackage[most]{tcolorbox}
\usepackage[normalem]{ulem} 

\newtcolorbox{memorycase}[1]{
    enhanced,
    breakable,
    colback=gray!4,
    colframe=gray!45,
    colbacktitle=gray!12,
    coltitle=black,
    fonttitle=\bfseries,
    title={#1},
    boxrule=0.5pt,
    arc=2pt,
    left=8pt,
    right=8pt,
    top=6pt,
    bottom=6pt,
    before skip=8pt,
    after skip=8pt,
    fontupper=\small
}

\usepackage{xcolor}
\usepackage{minitoc}
\usepackage{floatrow}
\tcbuselibrary{skins, breakable}

\newtcolorbox{promptbox}[2][]{
  enhanced,
  lines before break=4,          
  colback=white,
  colframe=black,
  boxrule=0.6pt,
  arc=2pt,
  left=6pt, right=6pt, top=4pt, bottom=4pt,
  title={\small\textbf{#2}},
  fonttitle=\small\bfseries,
  coltitle=black,
  attach boxed title to top left={yshift=-6pt, xshift=0pt},
  boxed title style={
    colback=white,
    colframe=black,
    boxrule=0.6pt,
    arc=2pt,
    left=4pt, right=4pt, top=1pt, bottom=1pt
  },
  toprule at break=0pt,
  bottomrule at break=0pt,
  pad at break=2pt,
  #1
}

\newcommand{\promptparam}[1]{%
  \texttt{\textcolor[gray]{0.3}{[}\textit{#1}\textcolor[gray]{0.3}{]}}%
}

\title{Multi-Head Recurrent Memory Agents}

\author{%
  Jiatong Li\quad Samuel Yeh\quad Sharon Li\\
  Department of Computer Science, University of Wisconsin-Madison\\
\texttt{\{jli2947, samuelyeh, sharonli\}@cs.wisc.edu}
}

\begin{document}
\doparttoc 
\faketableofcontents

\maketitle

\begin{abstract}
Recurrent memory agents extend LLMs to arbitrarily long contexts by iteratively consolidating input into a fixed-size memory window. Despite their scalability, these agents exhibit a well-documented reliability problem: end-to-end performance degrades systematically as context length grows. We diagnose this failure by decomposing performance into two factors---memory \emph{capture} and memory \emph{retention}---and quantitatively confirm that retention is the dominant bottleneck. Retention collapses because existing designs maintain memory as a monolithic text block, forcing every update to risk overwriting previously retained content. Motivated by this diagnosis, we propose \textbf{Multi-Head Recurrent Memory (MHM)}, a general, training-free framework that partitions memory into independent heads governed by a stage-wise select-then-update strategy. At each step, exactly one head is selected for update while the remaining heads are structurally shielded from overwriting, shifting the burden of retention from model behavior to architectural design. As a lightweight instantiation, we introduce Least-Recently-Updated MHM (MHM-LRU), which guarantees uniform head utilization with zero additional token overhead. Extensive experiments on long-context benchmarks show that MHM-LRU substantially improves both retention and end-to-end accuracy across the 100K--1M token range, where baselines degrade sharply. On RULER-HQA at 896K tokens, MHM-LRU improves the memory retention rate from less than 30\% to \textbf{73.96\%}. These gains generalize across model families, scales, and task types, positioning architectural optimization as a practical and cost-efficient path toward reliable long-context recurrent memory. Code: \url{https://github.com/deeplearning-wisc/multi-head-memory}
\end{abstract}

\section{Introduction}
The ability to reason over long contexts is a fundamental requirement for LLM agents operating
in the real world~\cite{zhang2024chain,jiang2024longllmlingua,li2023long,li2024loogle, xiao2024infllm}. Tasks such as document analysis~\cite{wang2024leave, ma2024mmlongbench, li2025long, wu2025resum}, multi-step research~\cite{huang2025deep, zhang2025web, chen2026iterresearch}, and extended dialogue~\cite{zhang2024chain, tan2025prospect}
demand that agents track and integrate information across inputs that far exceed the context
window. Recurrent memory~\cite{zhou2023recurrengpt, yuan2025memsearcher, zhou2026mem1} addresses this challenge by processing input
sequentially in chunks and iteratively compressing a long context into a fixed-size
memory window. At each step, the agent reads a new context chunk and updates its memory to integrate
new information with what has been retained so far. This paradigm is appealing because it places no hard limit
on context length and requires no modification to the underlying LLM. In practice, however,
recurrent memory agents exhibit a well-documented reliability problem: performance degrades
systematically as context length grows~\cite{yu2026memagent, du2025context}.

To motivate our work, we first diagnose {how} existing recurrent memory fails under long
contexts by decomposing end-to-end performance into two factors: whether the memory ever \emph{captures} the relevant information during memory update, and whether it \emph{retains} that information through all subsequent memory updates. Measuring both memory capture rate and memory retention rate on a representative method MemAgent~\cite{yu2026memagent}, we find that capture remains stable
across all context lengths, while memory retention decreases sharply, falling below 30\% at 896K
tokens. This diagnosis pinpoints memory
retention failure as the dominant bottleneck. The root cause is architectural: existing
recurrent memory designs maintain memory as a monolithic text block, so every update risks erasing key information. Information written
at step $t$ must survive $(T - t)$ subsequent overwrites to reach the final state, an
increasingly unlikely outcome as $T$ grows.

\begin{figure}
    \centering
    \includegraphics[width=\linewidth]{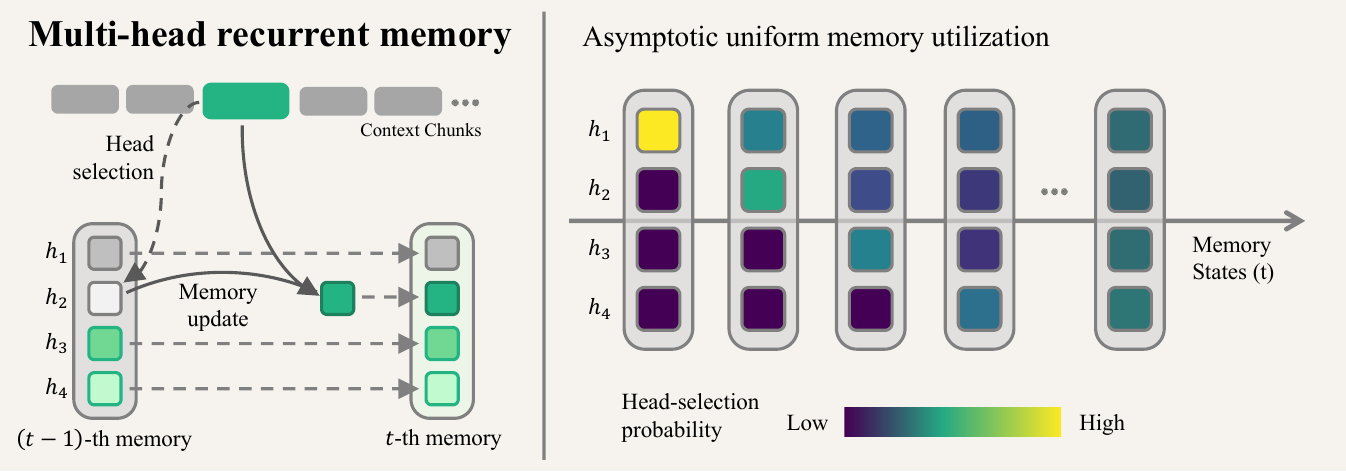}
    \caption{Left: An overview of the multi-head recurrent memory (MHM). Right: An overview of the memory head utilization pattern across time. In every time step, only one head is written. As time goes by, MHM achieves asymptotic uniform memory utilization.}
    \label{fig:diagnostic-framework}
\end{figure}

\par Motivated by the diagnosis, we propose \textbf{{Multi-Head Recurrent Memory (MHM)}}, a general, training-free memory framework designed to target memory retention failure. Rather than maintaining a single monolithic
memory block, MHM partitions memory into multiple independent heads and governs updates
through a stage-wise select-then-update strategy. At each step, one of the heads is selected
for update while the remaining heads are held fixed.
This
shifts the burden of retention from model behavior to the architectural design: rather than relying on the
model to avoid overwriting important content, 
MHM structurally shields the information retained in unselected heads from erasure at each step. The selection criterion is left general, accommodating any rule-based or
model-based strategy, and existing single-headed methods are subsumed as the special case. %further introduce \textbf{MHM-LRU}, a simple and effective 

As a lightweight instantiation, we showcase MHM-LRU, which selects the least-recently-updated head at each step. To diversify memory head use to persist memory retention, this rule provably guarantees uniform head utilization across heads and incurs zero additional token overhead relative to a single-headed baseline. Extensive experiments on realistic long-context benchmarks demonstrate that MHM-LRU substantially improves memory retention, outperforming baselines across all evaluated context lengths. Notably, on
RULER-HQA at 896K tokens, MHM-LRU improves memory retention rate from less than 30\% to \textbf{73.96}\%. The retention improvements directly translate into end-to-end performance gains, where MHM-LRU's accuracy remains strong and stable across the 100K-1M context range while baselines degrade sharply. These benefits generalize across model families, scales, and task types,
positioning architectural optimization as a practical and cost-efficient path toward reliable
long-context recurrent memory.
Overall, the contributions of this paper are summarized as follows:

\begin{enumerate}[leftmargin=*]
\item We identify memory retention failure as the dominant bottleneck of recurrent memory agents under extremely long contexts, supported by a quantitative capture-retention decomposition analysis.
\item We propose Multi-Head Recurrent Memory (MHM), a general and training-free
    memory architecture-level framework that structurally limits overwrite pressure through independent
    memory heads and stage-wise selective updates.
\item We extensively validate MHM  on realistic long-context benchmarks,
    demonstrating substantial retention and accuracy improvements across context lengths,
    model families, and task types. 
\end{enumerate}

\section{Preliminaries and Related Work}

\paragraph{Long-Context LLMs.} Scaling LLMs to long contexts has been pursued along two primary directions at the model level: extending the context window via positional encoding modifications~\cite{kazemnejad2023positional, he2024positional, wu2024positional} and reducing attention complexity via sparse or linear attention mechanisms~\cite{lai2025sparseattn, yang2025sparseattn, roy2021sparseattention}. These advances have enabled million-token-level context windows in both open-source models such as Qwen2.5-1M~\cite{yang2025qwen251m}, and proprietary systems such as Claude-Sonnet-4.6~\cite{anthropic2026claude} and Gemini-3.1-Pro~\cite{google2026gemini}.  Despite their effectiveness, native long-context LLMs remain bounded by fixed context limits or quadratic memory costs, motivating recurrent memory as a complementary paradigm that sidesteps both constraints.

\paragraph{Recurrent Memory Agents.} 

Early studies in agent memory~\citep{chhikara2025mem0,xu2026mem,packer2023memgpt} explore using an external memory store and retrieval to address long-term workflows. In contrast, recurrent memory appears to be targeted for single long-context processing of LLMs by recurrently maintaining a memory state. The idea of recurrent memory was foreshadowed in earlier architecture-level approaches~\cite{bulatov2022recurrent,gu2023mamba,peng2023rwkv,behrouz2024titans}. However, constrained by the representational capacity of the underlying architecture, the generality of these methods are limited in a broader range of applications. It was not until the emergence of token-level recurrent memory~\cite{yu2026memagent, zhou2023recurrengpt} that this paradigm was concretely established.

Formally, let $(C, q)$ denote a context-query pair where an LLM agent
must produce a correct response $y$. Since $C$ may far exceed the model's context window, the
agent partitions $C$ into an ordered chunk sequence $C = (c_1, c_2, \ldots, c_T)$ and maintains a
bounded memory state $m_t$ through sequential updates. The memory agent $f_\theta$ then processes these chunks sequentially, updating a bounded memory state $m_t$ at each step:
\begin{equation}
    m_t \leftarrow f_\theta(m_{t-1},\, c_t,\, q), \quad t = 1, 2, \ldots, T.
    \label{eq:memory-update}
\end{equation}
After all chunks have been processed, the LLM agent generates its final response conditioned solely on the accumulated memory:
\begin{equation}
    y \leftarrow \text{LLM}(m_T,\, q).
    \label{eq:final-response}
\end{equation}

Because $m_T$ is the {sole} information carrier from the first $T$ chunks to the final
response, the fidelity of the update in Eq.~\eqref{eq:memory-update} under growing $T$ is the
central reliability challenge of this paradigm. This paradigm has been concretely established by MemAgent~\cite{yu2026memagent} for long-context query answering. Subsequent work has extended it in several directions: GRU-Mem~\cite{sheng2026grumem} introduces text-controlled gating for selective
updates, while ReMemR1~\cite{shi2026remem} augments recurrent memory with a revisitable retrieval component, yet the core recurrent working memory remains a monolithic block. 

\paragraph{Single-Headed Memory as a Structural Bottleneck.}
Across existing recurrent memory
designs~\cite{yu2026memagent, shi2026remem, sheng2026grumem}, the core recurrent memory state $m_t$ is maintained
as a single monolithic text block---a design we term \emph{single-head} memory. At every update
step, the memory agent must overwrite this block with content derived from $c_t$, creating an
inherent tension between retaining previously captured information and incorporating new context.
This tension compounds with context length: information captured at step $t$ must survive $(T - t)$
subsequent overwrites intact to reach $m_T$. Under sufficiently long contexts, compression into a
single representational slot makes information loss structurally inevitable. In
the next section, we rigorously characterize and measure this failure, identifying it as the binding constraint on long-context performance.

\section{Characterizing Recurrent Memory Failure}\label{sec:diagnosis}

To motivate our memory architectural design, we first analyze \emph{how} recurrent memory fails under long
contexts. Incentivized by the working memory~\cite{baddeley2012working} and memory forgetting detection~\cite{spear1971forgetting,eldridge2005dissociation} in neuroscience, we decompose end-to-end performance into two memory factors: \textit{capture} and \textit{retention}.  For a correct
response, the memory must (1) \emph{capture} the relevant information at some point during
iterative updates, and (2) \emph{retain} it through all subsequent updates until the final state $m_T$.
Failure at either stage produces a wrong answer, but the two modes need not be equally prevalent
in practice.

\vspace{0.3cm}
\begin{definition}[\textbf{Memory Capture Rate (MCR)}]
Let $\mathcal{D} = \{(C_i, q_i, y^*_i)\}_{i=1}^N$ be an evaluation dataset and
$\mathcal{M} = \{m_i\}_{i=1}^N$ the corresponding memory
trajectories. The $m_i= \{m_{i,1}, \ldots, m_{i,T}\}$ denotes the memory state trajectory, where $T$ is the total number of trajectory steps for the respective instance. We assess answer membership $\mathbbm{1}(y^*_i \in m_{i,t})$ of the ground-truth answer against the memory content at step $t$. MCR is defined as the fraction of queries for which the relevant information was present in memory at least once
during processing:
\begin{equation}
    \mathrm{MCR}(\mathcal{M}, \mathcal{D}) =
    \frac{1}{N} \sum_{i=1}^{N}
    \mathbbm{1}\!\left(y^*_i \in m_{i,t},\; \exists\, t \in [T]\right).
    \label{eq:mcr}
\end{equation}
\end{definition}

 \vspace{0.3cm}
\begin{definition}[\textbf{Memory Retention Rate (MRR)}]
Among queries where memory capture occurred, the fraction for which the captured information survived
through to the final memory state $m_T$:
\begin{equation}
    \mathrm{MRR}(\mathcal{M}, \mathcal{D}) =
    \frac{
        \sum_{i=1}^{N}
        \mathbbm{1}(y^*_i \in m_{i,t},\exists\, t\in [T-1])\cdot \mathbbm{1}(y^*_i \in m_{i,T}) 
    }{
        \sum_{i=1}^{N}
        \mathbbm{1}(y^*_i \in m_{i,t},\; \exists\, t \in [T])
    }.
    \label{eq:mrr}
\end{equation}
{A high MRR indicates that the memory agent is capable of retaining previously captured information when processing irrelevant context. }
\end{definition}

\begin{figure}[t]
    \centering
    \includegraphics[width=\linewidth]{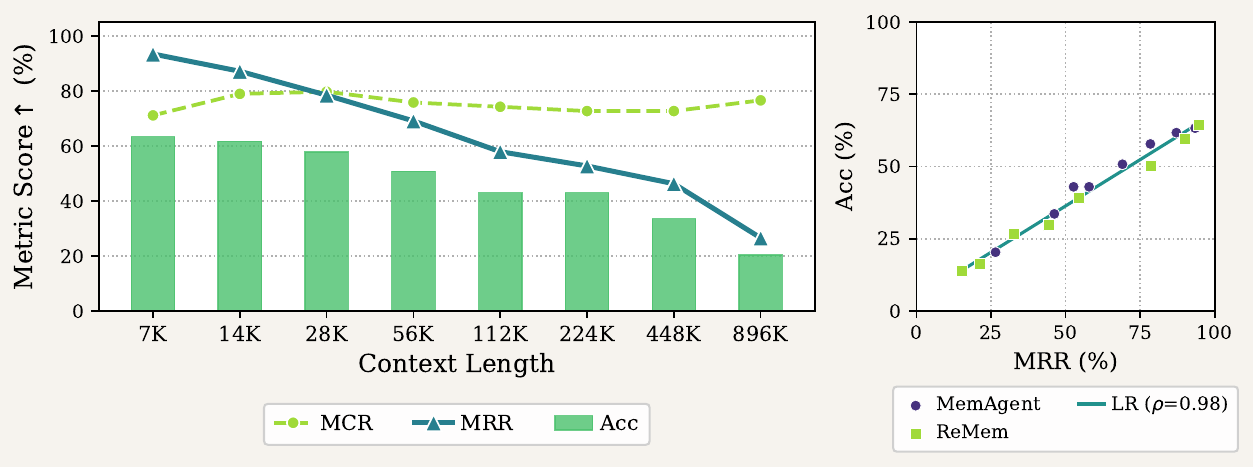}
    \caption{\small \textbf{(a)} MCR and MRR of MemAgent~\cite{yu2026memagent} (Qwen2.5-14B-Instruct) on RULER-HQA across context
lengths from 7K to 896K tokens. While MCR remains stable throughout, MRR
degrades sharply with context length, falling below 30\% at 896K tokens. \textbf{(b)}
MRR and Acc correlation analysis of MemAgent and ReMem~\cite{shi2026remem} on RULER-HQA across context lengths from 7K to 896K tokens. The `LR' denotes linear regression. End-to-end accuracy correlates strongly with MRR ($\rho = 0.98$), confirming that retention
failure is the dominant factor.}
    \label{fig:main}
\end{figure}
\paragraph{Retention Failure Dominates.}
We apply this analysis to MemAgent, a representative recurrent memory baseline built on  Qwen2.5-14B-Instruct~\cite{yang2024qwen25}, on the RULER-HQA~\cite{hsieh2024ruler,yang2018hotpotqa} benchmark. As shown in
Figure~\ref{fig:main} (left), MCR remains stable across all evaluated context lengths, indicating
that the agent consistently captures relevant information regardless of input length. MRR, by
contrast, degrades sharply, falling below 30\% at 896K tokens. Since MCR is roughly stable, the
observed accuracy degradation is almost entirely attributable to retention failure. Appending results from ReMem~\cite{shi2026remem}, Figure~\ref{fig:main} (right)
further confirms that end-to-end accuracy correlates strongly with MRR, with correlation coefficient $\rho = 0.98$. This isolates
memory retention as the main structural bottleneck, and directly motivates the design question we address
in Section~\ref{sec:method}: how to make the memory structure itself resistant to overwrite
pressure.

\section{Multi-Head Recurrent Memory}
\label{sec:method}

Motivated by our analysis, we propose \textbf{Multi-Head Recurrent Memory} (\textbf{MHM}), a general
framework designed to alleviate retention failure. The core insight is that retention failure is not a
failure of the model but the memory structure: when a single block holds all
context, every update is a potential erasure. MHM addresses this at the source by splitting memory into $H$ mutually independent {{memory heads}} and governing updates
through a {{selective overwrite}} strategy. More specifically, MHM represents the memory state as a set of $H$ independent text blocks,
$m_t = \{m^1_t, \ldots, m^H_t\}$, where each head $m^h_t$ is delimited by
\texttt{<memory\_\{h\}>~\ldots~</memory\_\{h\}>} tags. Each update affects only one memory head, while the remaining $(H - 1)$ heads are
structurally shielded from erasure by construction. Our design addresses two challenges: which head to
select for update, and how to perform the update given the selected head. We describe the two stages below.

\paragraph{Selection Stage.}
The selection stage addresses the memory retention failure through \textit{diversified} head selection---by spreading updates across different heads over time, the agent avoids repeatedly overwriting the same head and thereby reduces the risk of erasing pivotal information.
Formally, given the current memory state $m_{t-1} = \{m_{t-1}^1, \ldots, m_{t-1}^H\}$, the selection stage identifies the head $h^\star \in [H]$ most worthy of being updated:
\begin{equation}
    h^\star = \arg\max_{h \in [H]}\; r_{\mathrm{select}}(h \mid m_{t-1}, q, c_t).
    \label{eq:selection}
\end{equation}
where $r_{\text{select}}(h \mid m_{t-1}, q, c_t)$ is a criterion function that assigns 
a priority score to each head candidate. 
The criterion $r_{\mathrm{select}}$ can
encode a wide range of selection strategies depending on the available information and
computational budget. A rule-based criterion may select based on head age or a
fixed schedule. A model-based criterion may assess the relevance of each head given the current chunk and query. Regardless of the specific
instantiation, a well-designed criterion should satisfy a key property: \emph{diversity}, ensuring
that updates are spread across heads over time so that no single head bears disproportionate
overwrite pressure. We will provide an ablation on different selection strategies in Section~\ref{sec:ablation}.
\paragraph{Update Stage.}
The update stage overwrites the selected head $h^\star$ with content derived from the current context chunk.
Formally, given the current memory state $m_{t-1}$, the selected head index $h^\star$, the query $q$, and the current chunk $c_t$, the memory agent generates new head content and updates the memory state as follows:
\begin{equation}
    m^{h^\star}_t = f_\theta(m_{t-1},\, h^\star,\, q,\, c_t), \qquad
    m^h_t = m^h_{t-1} \;\; \forall\, h \neq h^\star.
    \label{eq:update}
\end{equation}
By receiving the full memory state $m_{t-1}$ as context, the update allows the memory agent to
reason jointly about what has already been retained across all heads and what new information
from $c_t$ is worth writing. This encourages both capturing newly relevant content and avoiding
redundancy with information already preserved in unselected heads. This stage-wise design subsumes existing single-headed methods as the special case $H = 1$, making MHM a flexible and general framework.%By dedicating a focused step solely to the update operation, the agent can attend fully to

\paragraph{A Lightweight Instantiation.}
As a concrete instantiation of MHM, inspired by the least-recently-used cache replacement algorithm~\cite{lee1999existence,jelenkovic2004least, megiddo2004outperforming}, we adopt a least-recently-updated (LRU) selection rule:
at each step, the head that has gone the longest without an update is selected. This instantiation has two advantages. First, the selection strategy ensures diversified head selection. When all heads
are initialized identically, this provides a
structural guarantee that no head is updated more than once before all others are updated at
least once. This ensures uniform utilization across heads and maximizes memory redundancy. Second, this method is computationally lightweight since head selection requires only a simple age comparison rather than an additional LLM inference pass, incurring
\emph{no extra token overhead} relative to a single-headed baseline. We refer to this
instantiation as \textbf{MHM-LRU}. As our experiments demonstrate, the structural guarantee of
uniform head utilization can substantially improve retention and end-to-end
performance in long context scenarios, suggesting that the benefits of MHM do not
depend on sophisticated selection strategies. Full prompt templates and implementation details
are provided in Appendix~\ref{app:mhm-lru}.

\section{Experiments}\label{sec:experiment}

\subsection{Experimental Setup}
\paragraph{Baselines.}
We compare MHM-LRU against two latest recurrent memory baselines: (1)~{MemAgent}~\cite{yu2026memagent}, which maintains a single memory head and applies a full overwrite at every update step, and (2)~{ReMem}~\cite{shi2026remem}, which maintains two heterogeneous heads (an update memory and a retrieval memory) where the retrieval memory is refreshed via a rule-based callback query.
GRU-Mem~\cite{sheng2026grumem} is excluded as its implementation is not publicly available.
All methods use Qwen2.5-14B-Instruct as the backbone of the recurrent memory structure. We additionally evaluate on Qwen2.5-32B-Instruct and gpt-oss-120b in the ablation study.
To ensure fair comparison, \emph{\textbf{the total memory capacity is kept the same across all methods}}: 4,096 tokens for MemAgent (one head), 2,048 per head for ReMem (two heads), and 1,024 per head
for MHM-LRU (4 heads). Full decoding and evaluation settings are provided in
Appendix~\ref{app:experiment-setting}.

\paragraph{Datasets.}
We evaluate on two benchmarks: (1)~
\texttt{RULER-HQA}~\cite{yu2026memagent},  a multi-hop QA dataset spanning context 
lengths from 7K to 892K tokens, following the evaluation protocol of~\cite{yu2026memagent}, and (2)~
\texttt{BABILong}~\cite{kuratov2024babilong}, a long-context reasoning benchmark covering 
ten structurally distinct task types (Q1--Q10) across context lengths from 1K to 10M tokens.
For BABILong, we evaluate on context lengths from 8K to 1M tokens. For each length, we sample 128 entries via systematic sampling to ensure 
uniform coverage of all task types. All recurrent memory methods are run three
times independently; we report the mean and standard deviation throughout.

\subsection{Main Results}

\begin{figure}[t]
    \centering
    \includegraphics[width=\linewidth]{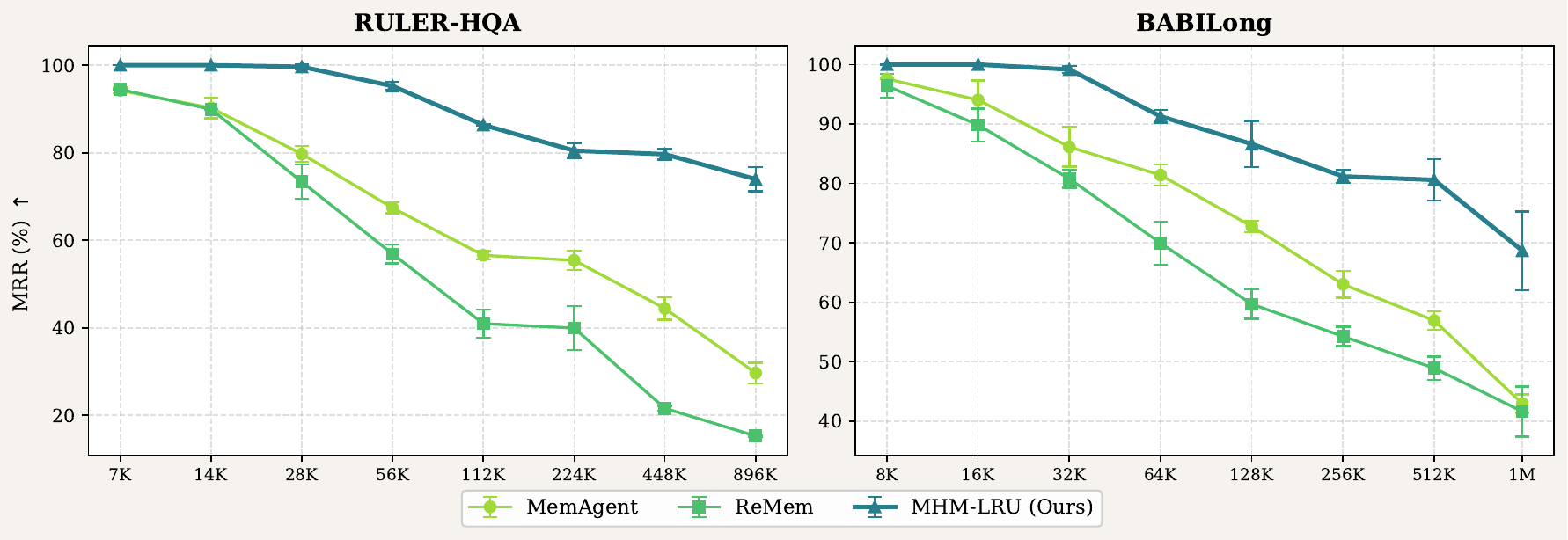}
    \caption{MRR results across different context lengths. The $x$-axis represents context length. Each data point is an average across 3 random runs with standard deviation as the error bar.}
    \label{fig:mdr-plot}
\end{figure}

\paragraph{MHM Substantially Improves Memory Retention}Figure~\ref{fig:mdr-plot} shows MRR across context lengths for all methods. We observe that MHM-LRU consistently
and substantially outperforms both baselines, with the gap widening at longer contexts. For example, on
RULER-HQA at 896K tokens, MHM-LRU achieves an MRR of \textbf{73.96}\% compared to less than 30\% for
MemAgent and ReMem. On BABILong at 1M tokens, MHM-LRU reaches 68.68\% versus
42.96\% for MemAgent. These results demonstrate the effectiveness of MHM-LRU in alleviating the memory retention issue of recurrent memories as context length increases. Moreover, at shorter context lengths (under 50K tokens), MHM-LRU achieves
near-perfect retention (MRR $>$ 99\%) on both datasets, confirming that the multi-head design
incurs no penalty in the common short-to-mid-length regime. Overall, the experiments demonstrate the effectiveness of MHM in addressing the retention of current memory.

\begin{table}[t]
\centering\footnotesize
\caption{End-to-end answer accuracy$\uparrow$ (\%) results on RULER-HQA.}\label{tab:acc-ruler-hqa}
\begin{tabularx}{\linewidth}{p{1.7cm}CCCCCCCC}
\toprule
\multirow{2}{*}{\textbf{Method}} & \multicolumn{8}{c}{\textbf{Context Length}} \\ \cmidrule(l){2-9}
 & \textbf{7K} & \textbf{14K} & \textbf{28K} & \textbf{56K} & \textbf{112K} & \textbf{224K} & \textbf{448K} & \textbf{896K} \\
 \midrule
LLM Baseline & \textbf{68.75} & \textbf{65.62} & 56.25 & \textbf{56.25} & 46.88 & 19.53 & 10.94 & 3.91 \\
MemAgent & 62.24\tiny{$\pm$0.97} & 60.94\tiny{$\pm$1.10} & 56.25\tiny{$\pm$2.21} & 50.52\tiny{$\pm$0.37} & 42.19\tiny{$\pm$0.64} & 42.97\tiny{$\pm$0.64} & 33.33\tiny{$\pm$0.98} & 21.62\tiny{$\pm$0.98} \\
ReMem & 64.32\tiny{$\pm$3.51} & 59.64\tiny{$\pm$0.98} & 50.26\tiny{$\pm$3.68} & 39.06\tiny{$\pm$2.78} & 29.68\tiny{$\pm$2.78} & 26.56\tiny{$\pm$1.69} & 16.14\tiny{$\pm$1.33} &  13.80\tiny{$\pm$0.97}
 \\
\midrule
MHM-LRU & {66.86\tiny{$\pm$1.69}} & {62.76\tiny{$\pm$1.47}} & \textbf{59.64\tiny{$\pm$1.33}} & 55.21\tiny{$\pm$1.33} & \textbf{50.26\tiny{$\pm$1.47}} & \textbf{50.26\tiny{$\pm$2.57}} &  \textbf{48.70\tiny{$\pm$2.88}} & \textbf{49.74\tiny{$\pm$1.47}}\\
\bottomrule
\end{tabularx}
\end{table}
\begin{table}[htp]
\centering
\footnotesize
\caption{End-to-end answer accuracy$\uparrow$ (\%) results on BABILong.}\label{tab:acc-babilong}
\begin{tabularx}{\linewidth}{p{1.7cm}CCCCCCCC}
\toprule
\multirow{2}{*}{\textbf{Method}} & \multicolumn{8}{c}{\textbf{Context Length}} \\ \cmidrule(l){2-9}
 & \textbf{8K} & \textbf{16K} & \textbf{32K} & \textbf{64K} & \textbf{128K} & \textbf{256K} & \textbf{512K} & \textbf{1M} \\
 \midrule
LLM Baseline & \textbf{64.06} & 57.81 & 53.12 & \textbf{53.12} & \textbf{50.0} & 35.16 & 31.25 & 27.34 \\
MemAgent & 58.85\tiny{$\pm$2.66} & 55.99\tiny{$\pm$2.58} & 53.39\tiny{$\pm$2.42} & 47.14\tiny{$\pm$0.98} & 44.66\tiny{$\pm$1.64} & 38.02\tiny{$\pm$2.88} & 32.55\tiny{$\pm$1.61} & 25.26\tiny{$\pm$0.97} \\
ReMem & 61.46\tiny{$\pm$2.41} & 55.21\tiny{$\pm$2.24} & 45.57\tiny{$\pm$1.61} & 42.19\tiny{$\pm$3.99} & 34.64\tiny{$\pm$2.42} & 32.29\tiny{$\pm$1.61} & 29.50\tiny{$\pm$0.26} & 24.48\tiny{$\pm$0.97} \\
\midrule
MHM-LRU & {63.54\tiny{$\pm$1.60}} & \textbf{60.94\tiny{$\pm$2.92}} & \textbf{55.47\tiny{$\pm$1.27}} & {51.82\tiny{$\pm$0.97}} & {48.70\tiny{$\pm$3.27}} & \textbf{47.39\tiny{$\pm$1.84}} & \textbf{46.35\tiny{$\pm$2.66}} & \textbf{41.41\tiny{$\pm$1.91}} \\
\bottomrule
\end{tabularx}
\end{table}

\paragraph{Retention Gains Translate Directly to End-to-End Performance.}  Table~\ref{tab:acc-ruler-hqa} and Table~\ref{tab:acc-babilong} report end-to-end accuracy across all context
lengths. We observe that MHM-LRU consistently outperforms both recurrent memory baselines and the native
long-context LLM across nearly all settings. The advantage is most pronounced at extreme context
lengths: on RULER-HQA at 896K tokens, MHM-LRU achieves 49.74\% accuracy versus 21.62\% for
MemAgent and 3.91\% for the LLM baseline; on BABILong at 1M tokens, 41.41\% versus 25.26\%
for MemAgent.  Crucially, MHM-LRU's accuracy remains stable across the 100K--1M range (48.70\% -- 50.26\% on RULER-HQA; 41.41\% -- 48.70\% on BABILong), whereas MemAgent and ReMem
degrade sharply. Taken together with the MRR results in Figure~\ref{fig:mdr-plot}, these findings confirm that the retention
improvements from MHM directly translate into downstream performance gains.

\subsection{Ablation Studies on Key Design}\label{sec:ablation}
We ablate two design dimensions of MHM:  the number of heads and the memory selection strategy. All
ablations are conducted on RULER-HQA using Qwen2.5-14B-Instruct, with per-head capacity fixed
at 1,024 tokens.%Our proposed memory diagnostic framework is used for the fine-grained analysis, interpreting the effect of different designs from the aspect of memory capture and memory decay.

\begin{wrapfigure}{r}{0.4\textwidth}
\vspace{-0.4cm}
  \centering
  \includegraphics[width=\linewidth]{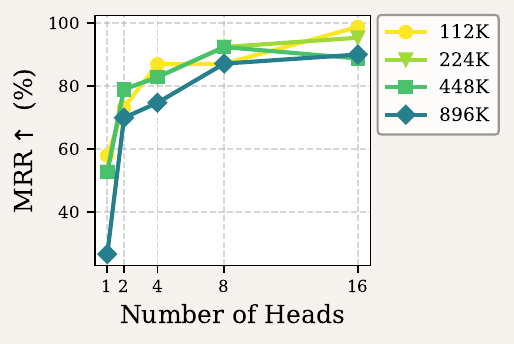}
  \vspace{-15pt}
  \caption{Effect of number of memory heads across context lengths.}
  \label{fig:ablation_heads}
\end{wrapfigure}
\paragraph{More Memory Heads Improve Retention.} We evaluate the impact of the number of heads $H$ on the memory retention rate (MRR). We vary $H$ from 1 to 16, under each context length. As shown in Figure~\ref{fig:ablation_heads}, MRR shows approximately logistic-like improvement as the number of heads grows. For example, under context length 896K, our approach improves MRR from 26.53\% with one head (equivalent to MemAgent) to 69.89\% with two heads, 74.67\% with four heads, and 87.12\% with eight heads, before plateauing at sixteen heads. This trend
holds consistently across all evaluated context lengths. This ablation study reveals that increasing the number of heads can significantly improve memory retention, confirming the structural advantage of our design.

\begin{wraptable}[19]{r}{0.45\linewidth}
\centering\footnotesize
\caption{Ablation study of the multi-headed memory selection strategy.}~\label{tab:ablation-selection}
\resizebox{\linewidth}{!}{
\begin{tabularx}{\linewidth}{lCCCC}
\toprule
 & \multicolumn{4}{c}{\textbf{Context Length}} \\
 \cmidrule(l){2-5}
\multirow{-2}{*}{\textbf{Metric}} & \textbf{112K} & \textbf{224K} & \textbf{448K} & \textbf{896K} \\
\midrule
\multicolumn{5}{l}{\cellcolor[HTML]{FAFAFA} \textit{MHM-Concur (w/o stage-wise)}} \\
Acc$\uparrow (\%)$ & 46.88 & 42.97 & 35.94 & 33.59 \\
MCR$\uparrow (\%)$ & {71.09} & {74.22} & {74.22} & 66.41 \\
MRR$\uparrow (\%)$ & {72.53} & 73.68 & 56.81 & 54.12 \\
\midrule
\multicolumn{5}{l}{\cellcolor[HTML]{F0F0F0} \textit{MHM-Relevance}} \\
Acc$\uparrow (\%)$ & 46.09 & {50.78} & 39.84 & 45.31 \\
MCR$\uparrow (\%)$ & {69.53} & {70.31} & {65.62} & {67.97} \\
MRR$\uparrow (\%)$ & {82.02} & {78.89} & {70.24} & {74.71} \\
\midrule
\multicolumn{5}{l}{\cellcolor[HTML]{E8F4EA} \textit{MHM-LRU}} \\
Acc$\uparrow (\%)$ & {50.26} & {50.26} & {48.70} & {49.74} \\
MCR$\uparrow (\%)$ & { 67.97} & { 71.88} & { 71.09} & { {69.16}} \\
MRR$\uparrow (\%)$ & {86.34} & {80.48} & {79.66} & {73.96} \\
\bottomrule
\end{tabularx}
}
\end{wraptable}

\paragraph{Stage-wise Selection Reduces Retention Failure.} 
A critical design in MHM is the stage-wise head selection and memory update. We thus compare MHM-LRU against a variant MHM-Concur to isolate the effect of the stage-wise design. In MHM-Concur, head selection and memory
update are merged into a single step: the model is asked to simultaneously decide which
head to overwrite and produce the updated content. 
As shown in Table~\ref{tab:ablation-selection}, while MHM-Concur achieves a notably higher MCR than the stage-wise approach, it suffers from substantially lower MRR (54.12\%
vs.\ 73.96\% at context length 896K), resulting in lower end-to-end accuracy across all context lengths. The
concurrent design introduces instruction complexity that degrades the model's ability to retain
captured content: when the model must simultaneously reason about head routing and content
generation, it is more likely to inadvertently erase previously retained information. In contrast, the stage-wise design eliminates this conflict by dedicating separate prompts to each operation,
allowing the update stage to focus entirely on capture and retention without the competing
objective of head routing. %Consequently, MHM-SLRU and MHM-LRU both achieve near-zero MDR at shorter context lengths (0.00\% at 112K--224K for MHM-SLRU), demonstrating that the stage-wise design is a critical factor in suppressing memory decay, even though it trades off some MCR relative to the concurrent approach.

\vspace{-0.2cm}
\paragraph{Rule-based vs. Relevance-based Selection.} 
We compare MHM-LRU (rule-based) against another variant, MHM-Relevance, to isolate
the effect of the selection criterion, holding the stage-wise design fixed. In MHM-Relevance, the
model is explicitly prompted to select the head most worthy of being updated, while diversifying the head usage (details in Appendix~\ref{app:mhm-relevance}). The intuition
is that the model can make relevance-aware routing decisions. Table~\ref{tab:ablation-selection} shows that, despite this appeal, MHM-Relevance usually incurs sub-optimal MCR and MRR, leading to lower end-to-end
accuracy (45.31\% vs. 49.74\% at context length 896K). The model-driven routing has two
compounding issues. First, model-based relevance assessment at each step is inherently noisy: the model
cannot reliably predict which information will be needed at query time, leading to
routing decisions that concentrate updates on heads perceived as salient and starve others of
updates. Second, this concentration undermines coverage diversity, reducing the breadth of
information captured across heads. MHM-LRU sidesteps the issues
 by guaranteeing uniform head
utilization without any model involvement, requires no additional inference for selection, and
achieves overall better performance. 

\subsection{Generality Analysis}
\label{sec:utility}

\begin{wraptable}{r}{0.55\linewidth}
\centering
\caption{\small MRR (\%) of different model families and scales. Best metric value given the model and context length is \textbf{bolded}.}\label{tab:mrr-model}
\resizebox{\linewidth}{!}{\begin{tabular}{llcccc}
\toprule
\multirow{2}{*}{\textbf{Model}} & \multirow{2}{*}{\textbf{Memory}} & \multicolumn{4}{c}{\textbf{Context Length}} \\
\cmidrule{3-6}
 &  & {112K} & {224K} & {448K} & {896K} \\
 \midrule
\multirow{2}{*}{gpt-oss-120b} & MemAgent & 75.00 & 55.84 & 61.64 & 35.71 \\
 & MHM-LRU & \textbf{84.48} & \textbf{75.44} & \textbf{69.35} & \textbf{68.85} \\
 \midrule
\multirow{2}{*}{Qwen2.5-32B} & MemAgent & 84.04 & 87.63 & 88.54 & 78.02 \\
 & MHM-LRU & \textbf{93.02} & \textbf{95.65} & \textbf{90.00} & \textbf{81.91} \\
 \bottomrule
\end{tabular}}
\end{wraptable}
\paragraph{Generality Across Model Family and Scale.}
We evaluate MHM-LRU with two additional base models: Qwen2.5-32B-Instruct
and gpt-oss-120b. We focus on the 112K--896K settings, where retention failure is most severe. Across both models, MHM-LRU maintains significantly higher and more stable
MRR than MemAgent, as shown in Table~\ref{tab:mrr-model}. On gpt-oss-120b, MemAgent's MRR falls below 40\%, while MHM-LRU maintains MRR to 68.85\%. On Qwen2.5-32B-Instruct, MHM-LRU's MRR consistently achieves 90\% on 112K--448K and maintains to 81.91\% on 896K.
This consistent
pattern across model families and scales confirms that the retention benefit of MHM does not
depend on any model-specific property, but follows from the memory architectural advantage.

\begin{figure}
    \begin{floatrow}
      \ffigbox[0.68\textwidth]{\caption{\small Accuracy by task type (QA1 -- QA10) and context length (128K -- 1M) on BABILong.}\label{fig:generality-task-acc}}{%
        \includegraphics[width=\linewidth]{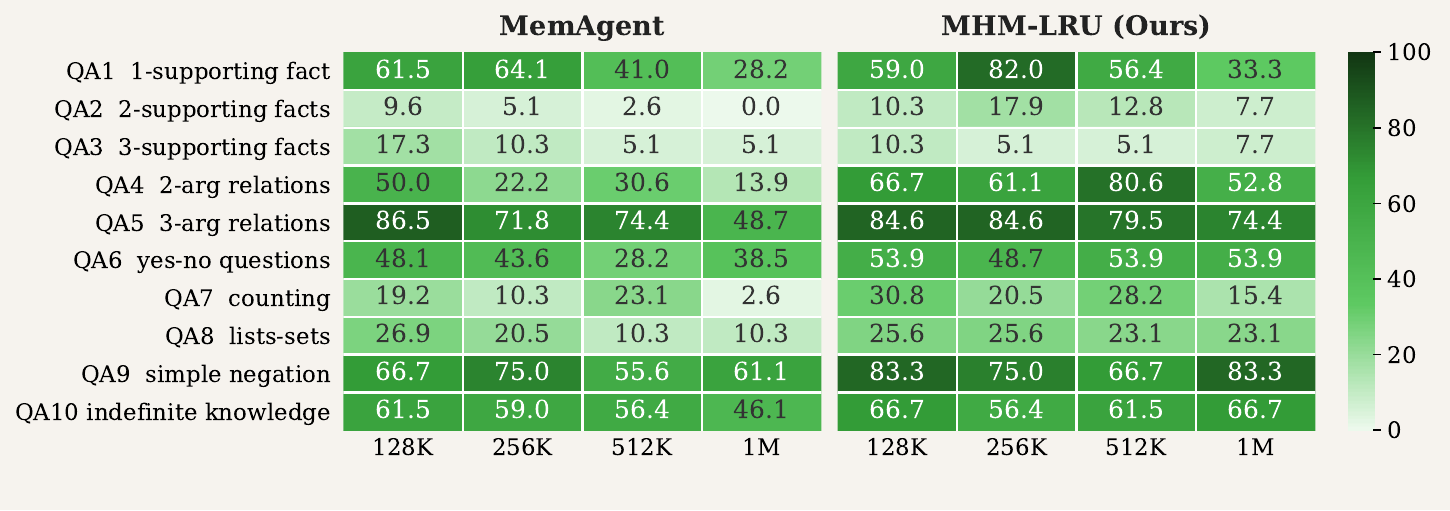}
      }
      \ffigbox[0.325\textwidth]{\caption{\small VRAM cost and accuracy change on RULER-HQA.}\label{fig:generality-pareto}}{%
        \includegraphics[width=\linewidth]{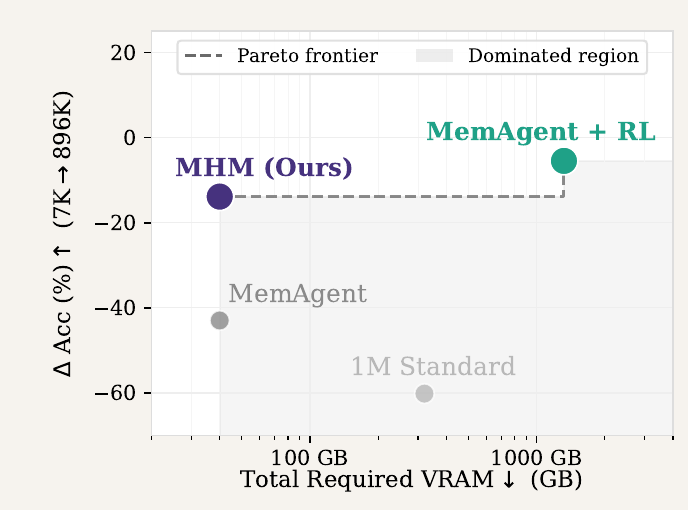}
      }
    \end{floatrow}
    \vspace{-0.5cm}
\end{figure}

\vspace{-0.2cm}
\paragraph{Generality Across Task Types.} To evaluate the generality of MHM-LRU across diverse task types, we conduct 
experiments on BABILong, which encompasses ten structurally distinct reasoning 
tasks (QA1--QA10). As shown in Figure~\ref{fig:generality-task-acc}, MHM-LRU outperforms MemAgent across all ten task types at long 
contexts (1M), with particularly pronounced gains on tasks requiring 
relational reasoning and yes-no judgment (QA4--QA6), where MemAgent's accuracy 
degrades sharply with context length. This pattern aligns with our retention analysis: tasks involving multi-step relational chains are most susceptible to 
memory retention failure, making them the
primary beneficiaries of MHM's structural redundancy. These 
results confirm that MHM-LRU's advantage is not task-specific but reflects 
a systematic improvement that generalizes across task types.

\begin{table}[htp]
\centering\footnotesize
\caption{Time and output token costs of recurrent memory agents on RULER-HQA. For each method, the test is run on an NVIDIA A100 80GB, with four parallel threads.}\label{tab:time-token-cost}
\begin{tabularx}{\linewidth}{p{2.4cm}p{1.5cm}CCCCCCCC}
\toprule
\multicolumn{2}{c}{\textbf{Metric | Method}} & \textbf{7K} & \textbf{14K} & \textbf{28K} & \textbf{56K} & \textbf{112K} & \textbf{224K} & \textbf{448K} & \textbf{896K} \\
\midrule
\multirow{2}{*}{\begin{tabular}[c]{@{}l@{}}Time / Sample (sec)\end{tabular}} & MemAgent  & 2.8 & 5.1 & 9.2 & 17.3 & 31.5 & 59.1 & 114.3 & 224.9 \\
 & MHM-LRU  & 1.9 & 3.3 & 6.5 & 13.0 & 25.9 & 53.4 & 117.1 & 239.1 \\
 \midrule
\multirow{2}{*}{Tokens / Sample} & MemAgent & 330.6 & 535.8 & 916.2 & 1705.6 & 2834.4 & 4943.4 & 9340.6 & 18013.9 \\
 & MHM-LRU & 142.8 & 214.4 & 419.5 & 898.9 & 1875.7 & 3814.0 & 9038.6 & 18085.5 \\
\bottomrule
\end{tabularx}
\end{table}

\vspace{-0.5cm}
\paragraph{Computational Efficiency.} 
Figure~\ref{fig:generality-pareto} plots accuracy degradation (7K to 896K) against total VRAM cost for
all methods. We additionally include MemAgent+RL (RL-MemAgent~\cite{yu2026memagent}),  a reinforcement learning post-trained variant of MemAgent that requires significantly more VRAM due to the overhead of RL training. MHM-LRU lies on the Pareto frontier between these two extremes: it recovers a substantial fraction of the accuracy gain achieved by MemAgent+RL at no additional memory overhead  beyond vanilla MemAgent. This is notable because MHM-LRU requires neither model weight access nor task-specific training, and the gain comes entirely from the architectural design with multiple heads. This positions architectural optimization as a practical and resource-efficient path toward reliable long-context recurrent memory, particularly in settings where model retraining or large VRAM budgets are unavailable.

\par We further evaluate the time 
and output token costs of different recurrent memory methods under long contexts, as shown in Table~\ref{tab:time-token-cost}. MHM-LRU 
is consistently faster than MemAgent at short-to-mid context lengths (e.g., 25.9s vs.\ 31.5s 
at 112K), with negligible overhead at extreme lengths. Similarly, MHM-LRU consumes 
substantially fewer output tokens at shorter contexts (142.8 vs.\ 330.6 at 7K), converging 
to parity with MemAgent only at 896K. Both effects reflect a spontaneous amortization 
property of MHM: by writing to a single head per step, the LLM naturally generates fewer 
tokens per update, amortizing the cost of memory maintainance. These results confirm that 
MHM-LRU achieves its retention and accuracy gains at no meaningful increase in 
computational cost.

\begin{figure}[t]
\vspace{-0.5cm}
    \centering
    \includegraphics[width=\linewidth]{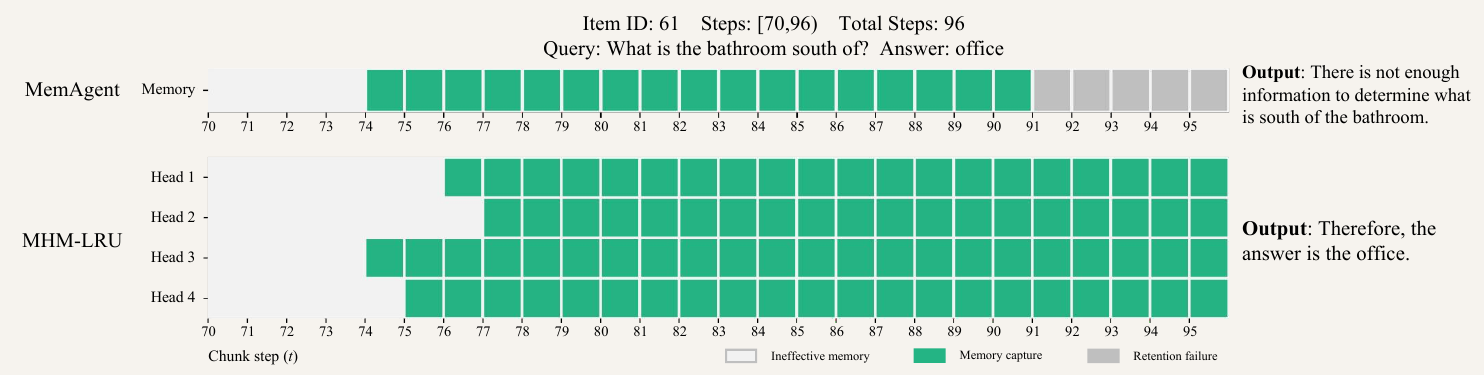}
    \caption{A case study of memory trajectory on BABILong-512K. The task type is QA4: 2-arg relations.}
    \label{fig:babilong_case}
    \vspace{-0.2cm}
\end{figure}

\vspace{-0.2cm}
\subsection{Memory Dynamics Case Study}\label{sec:case-study}
\vspace{-0.2cm}
To concretely illustrate how MHM-LRU alleviates retention failure, Figure~\ref{fig:babilong_case} presents a representative memory trajectory on a BABILong-512K sample from the 2-arg relations task. The query asks ``What is the bathroom south of?'' (ground truth: ``office''), with relevant chunks falling within steps [70, 96). MemAgent captures the answer from step 74 onward but overwrites it during the final updates (steps 91–95), yielding the incorrect response ``There is not enough information to determine what is south of the bathroom.'', which is a representative instance of retention failure. MHM-LRU, in contrast, captures the answer at step 74 and then spontaneously propagates it to the remaining heads in subsequent updates, maintaining multiple redundant copies that persist through step 95 and yielding the correct answer ``Therefore, the answer is the office.'' This emergent backup behavior concretely demonstrates how partitioning memory into independent heads converts the stochastic overwrite risk of a monolithic block into an architectural retention guarantee.

\vspace{-0.3cm}
\section{Conclusion and Limitation}\label{sec:conclusion}
\vspace{-0.3cm}
\par We presented Multi-Head Recurrent Memory (MHM), a new framework for improving the
reliability of recurrent memory agents under long contexts. Through a diagnostic
analysis of memory dynamics, we identified retention failure as the dominant bottleneck of
long-context performance.  Motivated by this diagnosis, MHM is designed to address the memory retention failure through multiple memory heads and the stage-wise memory update strategy. MHM  structurally limits overwrite pressure on retained
content without requiring any model retraining. 
MHM-LRU, a lightweight instantiation, substantially improves memory retention and
end-to-end accuracy on challenging long-context query benchmarks. We hope these findings
encourage future work on architectural approaches to long-context reliability, and on combining
MHM with training-based methods for further gains.
\vspace{-0.2cm}
\paragraph{Limitation.} In our experiments, the number of heads $H$ is fixed. While our ablation shows that
MRR improves monotonically with $H$, the optimal number of heads is likely dependent on
context length and task structure.
An adaptive head allocation strategy that adjusts $H$ based on input characteristics is a
natural direction for future work. 

\section*{Acknowledgment}
We sincerely thank Changdae Oh, Linyue Cai and Hyeong Kyu Choi for proofreading the manuscript, and Jimmy Di for providing additional computational resources. This work is supported in part by the AFOSR Young Investigator Program under award number FA9550-23-1-0184, Office of Naval Research under award N000142612508, National Science Foundation under awards IIS-2237037 and IIS-2331669,  Schmidt Sciences Foundation, Open Philanthropy (now Coefficient Giving), Alfred P. Sloan Fellowship, and gifts from Google and Amazon.
 
\bibliographystyle{unsrt}
\bibliography{custom}

\newpage
\appendix
\begingroup
\renewcommand{\thepart}{}       
\renewcommand{\partname}{}     
\part{Appendix}                 
\parttoc
\endgroup                       
\section{Multi-Head Recurrent Memory Paradigm}

\subsection{Overview of MHM Pipeline}\label{app:mhm-overview}
\par The pseudo-code of the MHM pipeline is shown in Algorithm~\ref{alg:mhm}.

\begin{algorithm}
\caption{Multi-Headed Memory For Long-Context Query}\label{alg:mhm}
\KwIn{Context $C$, query $q$, backbone LLM, number of heads $H$, chunk size $L$}
\KwOut{Response $y$}

Initialize memory $m_0 \leftarrow \{m_0^1, \ldots,m_0^H|m_0^h=\emptyset, h\in[H]\}$\;
Equally split $C$ into context chunks $C_{\text{split}}\leftarrow[c_1,c_2,\ldots,c_T], T=\left\lceil \frac{|C|}{L}\right\rceil$.

\For{$t \leftarrow 1$ \KwTo $T$}{
    Head selection: $h^\star \leftarrow \arg\max_{h' \in [H]}\, r_{\text{select}}(h' \mid m_{t-1}, q, c_t)$\;
    Head update: $m_t^{h^\star} \leftarrow \text{LLM}(m_{t-1},\, h^\star,\, q,\, c_t)$\;
    Memory Overwrite: $m_{t} = \left(m_{t-1} \setminus \{m_{t-1}^{h^\star}\}\right) \cup \{m_t^{h^\star}\}$.
}
Generate response $y \leftarrow \pi_\theta(m_T,\, q)$\;

\Return $y$\; 
\end{algorithm}

\subsection{MHM-LRU: MHM Instantiation with Least Recently Updated Strategy}\label{app:mhm-lru}
\par The prompt template of MHM-LRU is shown in Figure~\ref{fig:prompt-template-mhm-lru}.

\begin{figure}[htp]
\begin{promptbox}{Prompt Template: Stage-wise Memory Head Update and Final Answer}

\medskip
\textbf{Prompt Template for Memory Update:}

\medskip
You are presented with a problem, a section of an article that may contain the answer to the
problem, and previous memory. Please read the provided section carefully and update the
memory \textbf{*\promptparam{memory ID}*} with the new information that helps to answer
the problem. Be sure to retain all relevant details from the previous memory while adding
any new, useful information. Directly output the updated content without including the
memory id. Be sure to note the source context location of the updated memory.

\medskip
\texttt{<problem>}\\
\promptparam{Query $q$}\\
\texttt{</problem>}

\begin{minipage}[htp]{0.3\linewidth}
\medskip
\texttt{<memory\_1>}\\
\promptparam{Memory Head $m_{t-1}^{1}$}\\
\texttt{</memory\_1>}

\medskip
\texttt{<memory\_2>}\\
\promptparam{Memory Head $m_{t-1}^{2}$}\\
\texttt{</memory\_2>}

\medskip
\texttt{<memory\_3>}\\
\promptparam{Memory Head $m_{t-1}^{3}$}\\
\texttt{</memory\_3>}

\medskip
\texttt{<memory\_4>}\\
\promptparam{Memory Head $m_{t-1}^{4}$}\\
\texttt{</memory\_4>}
\end{minipage}%
\begin{minipage}[t]{0.05\linewidth}
\textcolor{gray}{$\left.\rule{0pt}{8em}\right\}$}
\end{minipage}%
\begin{minipage}[t]{0.7\linewidth}
\promptparam{Memory Head Set $m_{t-1}=\{m_{t-1}^1,m_{t-1}^2,m_{t-1}^3,m_{t-1}^4\}$}
\end{minipage}

\medskip
\texttt{<section>}\\
\promptparam{Context Chunk $c_t$}\\
\texttt{</section>}

\medskip
Updated \promptparam{memory ID}:

\noindent\textcolor{gray!50}{\rule{\linewidth}{0.4pt}}\\
\textbf{Prompt Template for Final Answer:}

\medskip
You are presented with a problem and a previous memory containing multiple heads. Please answer the problem based on the previous memory and format your response as follows "Therefore, the answer is (insert answer here)".

\medskip
\texttt{<problem>}\\
\promptparam{Query $q$}\\
\texttt{</problem>}

\medskip
\promptparam{Memory Head Set $m_T=\{m_T^1,m_T^2,m_T^3,m_T^4\}$}

\medskip
Your answer:

\end{promptbox}
\caption{LLM prompt template used in MHM-LRU and MHM-Relevance. Italicized tokens in brackets denote runtime input parameters. It should be noticed that \textbf{the targeted memory head is selected and input as a parameter by the previous selection stage}.}\label{fig:prompt-template-mhm-lru}
\end{figure}

\subsection{Theoretical Guarantee of MHM-LRU}
\label{appendix:lru_proof}

We formally establish the uniform head utilization guarantee of MHM-LRU claimed in 
Section~\ref{sec:method}. Let $n_h(t)$ denote the number of times head $h \in [H]$ 
has been selected for update by the end of step $t$. Note that $\sum_{h=1}^{H} n_h(t) = t$ 
for all $t \geq 0$.

\begin{proposition}[Uniform Head Utilization]
\label{prop:lru}
Under MHM-LRU with $H$ heads all initialized identically, for all $t \geq 0$:
\begin{equation}
    \max_{h \in [H]}\, n_h(t) \;-\; \min_{h \in [H]}\, n_h(t) \;\leq\; 1.
\end{equation}
Equivalently, every head satisfies
\begin{equation}
    \left\lfloor \frac{t}{H} \right\rfloor \;\leq\; n_h(t) \;\leq\; 
    \left\lceil \frac{t}{H} \right\rceil, \qquad \forall\, h \in [H].
\end{equation}
\end{proposition}

\begin{proof}
We proceed by induction on $t$.

\textbf{Base case} ($t = 0$): $n_h(0) = 0$ for all $h$, so $\max_h n_h(0) - \min_h n_h(0) = 0 \leq 1$.

\textbf{Inductive step}: Assume that at step $t$, $\max_h n_h(t) - \min_h n_h(t) \leq 1$.
Let $n_{\min} \coloneqq \min_h n_h(t)$.  By the inductive hypothesis, every head satisfies 
$n_h(t) \in \{n_{\min},\, n_{\min}+1\}$.

MHM-LRU selects $h^{\star} = \arg\max_{h \in [H]}\, r_{\text{select}}(h \mid m_{t-1}, q, c_t) = \arg\min_{h} n_h(t)$, so $n_{h^\star}(t) = n_{\min}$.
After the update:
\[
    n_h(t+1) = \begin{cases} n_{\min} + 1 & h = h^\star, \\ n_h(t) & h \neq h^\star. \end{cases}
\]
We consider two sub-cases.

\textit{Sub-case A}: All heads share count $n_{\min}$ at step $t$.  After the update,
$n_{h^\star}(t+1) = n_{\min}+1$ and $n_h(t+1) = n_{\min}$ for all $h \neq h^\star$.
Hence $\max_h n_h(t+1) - \min_h n_h(t+1) = 1 \leq 1$.

\textit{Sub-case B}: At least one head other than $h^\star$ has count $n_{\min}+1$ at 
step $t$.

\begin{itemize}[leftmargin=*]
    \item If there exists $h' \neq h^\star$ with $n_{h'}(t) = n_{\min}$, then after the update 
    $\min_h n_h(t+1) = n_{\min}$ and $\max_h n_h(t+1) = n_{\min}+1$, giving 
    $\max - \min = 1 \leq 1$.
    \item If $h^\star$ is the \emph{unique} head with count $n_{\min}$ at step $t$, then all other 
    heads already hold count $n_{\min}+1$.  After the update every head has count $n_{\min}+1$, 
    so $\max - \min = 0 \leq 1$.
\end{itemize}

In all cases the invariant is preserved.  By induction, the bound holds for all $t \geq 0$.

To verify the floor–ceiling form: from the invariant $n_h(t) \in \{n_{\min}, n_{\min}+1\}$ 
and the identity $\sum_h n_h(t) = t$, we have 
$H \cdot n_{\min} \leq t \leq H \cdot (n_{\min}+1)$, i.e.,
$n_{\min} = \lfloor t/H \rfloor$.  
Hence $\lfloor t/H \rfloor \leq n_h(t) \leq \lceil t/H \rceil$ for all $h$.
\end{proof}

\newtheorem{corollary}{Corollary}

\begin{corollary}[Asymptotic Uniform Utilization]
\label{cor:asymptotic}
Under MHM-LRU, the per-head update fraction converges uniformly to $1/H$:
\begin{equation}
    \left|\frac{n_h(t)}{t} - \frac{1}{H}\right| \;\leq\; \frac{1}{t}, 
    \qquad \forall\, h \in [H],\; t \geq 1.
\end{equation}
In particular, $n_h(t)/t \to 1/H$ as $t \to \infty$, uniformly over all heads.
\end{corollary}

\begin{proof}
From Proposition~\ref{prop:lru}, $\lfloor t/H \rfloor \leq n_h(t) \leq \lceil t/H \rceil$.  
Since $\lceil t/H \rceil - t/H \leq 1$ and $t/H - \lfloor t/H \rfloor \leq 1$, we have 
$|n_h(t) - t/H| \leq 1$, which gives $|n_h(t)/t - 1/H| \leq 1/t$.
\end{proof}

\paragraph{Remark.} Proposition~\ref{prop:lru} implies that no head bears disproportionate 
overwrite pressure: across any $H$ consecutive steps, every head is updated \emph{exactly} 
once.  This structural guarantee is the formal basis for the retention advantage of MHM-LRU 
observed in Section~\ref{sec:experiment}: by bounding the maximum number of overwrites any 
single head sustains, the probability that a previously captured answer survives to the final 
memory state is maximized uniformly across all heads.  Crucially, this guarantee is 
\emph{architectural} rather than behavioral. This guarantee holds regardless of the LLM backbone or task, and requires no additional inference overhead, since head selection reduces to a simple 
integer comparison.

\subsection{MHM-Concur: MHM Instantiation with Concurrent Selection And Update}
\par MHM-Concur is a vanilla instantiation of MHM. The motivation of MHM-Concur is that the memory agent should be capable of selecting and updating the memory head at the same time. As shown in Figure~\ref{fig:prompt-template-mhm-concur}, in the memory update prompt template, the LLM is required to output the updated memory in a specific format that consists of the selected head ID and updated content. The final answer prompt template is the same as in Figure~\ref{fig:prompt-template-mhm-lru}.

\begin{figure}[htp]
\begin{promptbox}{Prompt Template: Concurrent Memory Head Update}
\medskip
You are presented with a problem, a section of an article that may contain the answer to the problem, and previous memory. Please read the provided section carefully and update the memory with the new information that helps to answer the problem. Specifically, the memory consists of four memory heads: \texttt{memory\_1, memory\_2, memory\_3, memory\_4}. Please select one of the memory heads and update the memory. Be sure to diversify the use of different memory heads within \texttt{memory\_1} to \texttt{memory\_4} and avoid repeatedly selecting the same memory head for updates. Be sure to retain all relevant details from the previous memory while adding any new, useful information. Be sure to format your update as follows: "\texttt{<memory\_id> (your updated memory content) </memory\_id>}".

\medskip
\texttt{<problem>}\\
\promptparam{Query $q$}\\
\texttt{</problem>}

\medskip
\promptparam{Memory Head Set $m_{t-1}=\{m_{t-1}^1,m_{t-1}^2,m_{t-1}^3,m_{t-1}^4\}$}

\medskip
\texttt{<section>}\\
\promptparam{Context Chunk $c_t$}\\
\texttt{</section>}

\medskip
Updated memory:
\end{promptbox}
\caption{LLM prompt template for memory update in MHM-Concur. Italicized tokens in brackets denote runtime input parameters. Different from the stage-wise paradigm, MHM-Concur requires the LLM to \textbf{select and update memory heads at the same time}.}
\label{fig:prompt-template-mhm-concur}
\end{figure}

\subsection{MHM-Relevance: MHM Instantiation with Model-Based Selection Strategy}\label{app:mhm-relevance}

\par MHM-Relevance is characterized by the stage-wise, model-driven, semantic-aware head selection --- another important potential instantiation of MHM. Specifically, the selection criteria is explicitly regulated as selecting the head that is the most semantically similar to the section, while prioritizing selecting empty memory heads to avoid single-head memory pattern, as shown in Figure~\ref{fig:prompt-template-mhm-slru}. The consequential memory update and final answer prompt templates are the same as MHM-LRU, as shown in Figure~\ref{fig:prompt-template-mhm-lru}.

\begin{figure}[htp]
\begin{promptbox}{Prompt Template: Model-Based Relevance-Driven Selection}
\medskip
You are presented with a problem, a section of an article that may contain the answer to the problem, and previous memory. Please read the provided section carefully and select a memory head in \texttt{memory\_1, memory\_2, memory\_3, memory\_4} that is *\textbf{the most semantically similar to the section}*. If there is an empty memory head, you should prioritize selecting it. Be sure to diversify the use of different memory heads within \texttt{memory\_1} to \texttt{memory\_4} and avoid repeatedly selecting the same memory head for updates. 

\medskip
\texttt{<problem>}\\
\promptparam{Query $q$}\\
\texttt{</problem>}

\medskip
\texttt{<memory\_1>}\\
\promptparam{Memory Head $m_{t-1}^{1}$}\\
\texttt{</memory\_1>}

\medskip
\texttt{<memory\_2>}\\
\promptparam{Memory Head $m_{t-1}^{2}$}\\
\texttt{</memory\_2>}

\medskip
\texttt{<memory\_3>}\\
\promptparam{Memory Head $m_{t-1}^{3}$}\\
\texttt{</memory\_3>}

\medskip
\texttt{<memory\_4>}\\
\promptparam{Memory Head $m_{t-1}^{4}$}\\
\texttt{</memory\_4>}

\medskip
\texttt{<section>}\\
\promptparam{Context Chunk $c_t$}\\
\texttt{</section>}

\medskip
Selected memory:
\end{promptbox}
\caption{LLM prompt template for memory selection in MHM-Relevance. Italicized tokens in brackets denote runtime input parameters. }
\label{fig:prompt-template-mhm-slru}
\end{figure}
\section{Detailed Experiment Setting}\label{app:experiment-setting}
\subsection{Memory Setting}
\paragraph{Memory Setting.} The context chunk size is set to 5,000 tokens per chunk for long context splitting, consistent with the memory setting in~\cite{yu2026memagent}. The total memory capacity is set to 4,096 tokens for all recurrent memory agents, which requires setting the token generation limit for each method according to its number of heads. 
\begin{itemize}[leftmargin=*]
    \item \textbf{MemAgent} (one single memory tag): The \texttt{max\_new\_tokens} is set to 4,096.
    \item \textbf{ReMem} (one current memory and one recalled memory tags): The \texttt{max\_new\_tokens} is set to 2,048, ensuring that the total capacity of the two memory heads equals 4,096.
    \item \textbf{MHM-LRU}: The number of heads is set to four, and the \texttt{max\_new\_tokens} is set to 1,024, ensuring that the total capacity of the four memory heads equals 4,096. At the beginning, when multiple memory heads have the same highest age, MHM-LRU chooses the lowest available head index by default.
    \item \textbf{Ablation on \# of heads for MHM-LRU}: The \texttt{max\_new\_tokens} is fixed to 1,024 regardless of the number of heads, ensuring that the agent's performance is not affected by the change of token generation limit.
\end{itemize}

\subsection{Runtime Setting}
\paragraph{Decoding Setting.} In the LLM decoding setting, the \texttt{temperature} is set as 0.7. The \texttt{top\_p} is set as 0.95. The timeout limit is set as 60 seconds per 1,024 output tokens. As a result, the timeout limit is 240 seconds for MemAgent, 120 seconds for ReMem, and 60 seconds for MHM-LRU. Each LLM calling has three tries in case of encountering an exception (e.g., a timeout exception). All experiments are run on a Linux server with 8$\times$ A100 GPUs. LLMs are deployed with vLLM 0.15.0.
\paragraph{Evaluation Setting.} In evaluating the end-to-end performance (accuracy) of recurrent memory agents, we adopt substring exact matching. In the model's final response, we instruct the model to output the answer in the format ``\texttt{Therefore, the answer is (insert answer here)}''. The evaluation pipeline matches the output sentence starting with ``\texttt{Therefore, the answer is }'', with the rest part is distinguished as the agent's prediction. The pipeline then executes substring exact matching between normalized ground truth and the agent's prediction to decide the accuracy score of the agent's final response.
\section{Additional Experiment Results}
\subsection{Ablation On Head Count Fixing Total Capacity}

\begin{table}[htp]
    \caption{Memory Retention Rate (MRR $\uparrow$ \%) with respect to head count given fixed total memory capacity.}\label{tab:mrr-fix-total}
\resizebox{0.55\linewidth}{!}{\begin{tabular}{lrrrrr}
\toprule
\multirow{2}{*}{\textbf{Length}} & \multicolumn{5}{c}{\textbf{Head Count}} \\
\cmidrule{2-6}
 & 1 & 2 & 4 & 8 & 16 \\ \midrule
112K & 57.89 & 89.77 & 86.34 & 94.51 & 100.00 \\ 
224K & 52.69 & 84.69 & 80.48 & 93.48 & 98.86 \\ 
448K & 46.24 & 78.72 & 79.66 & 89.25 & 91.40 \\ 
896K & 26.53 & 80.68 & 73.96 & 87.50 & 96.25 \\
\bottomrule
\end{tabular}}
\end{table}

\begin{table}[htp]
    \caption{Accuracy (Acc $\uparrow$ \%) with respect to head count given fixed total memory capacity.}\label{tab:acc-fix-total}
\resizebox{0.55\linewidth}{!}{
\begin{tabular}{lrrrrr}
\toprule
\multirow{2}{*}{\textbf{Length}} & \multicolumn{5}{c}{\textbf{Head Count}} \\
\cmidrule{2-6}
 & 1 & 2 & 4 & 8 & 16 \\ \midrule
112K & 42.19 & 54.69 & 50.26 & 58.59 & 61.72 \\ 
224K & 42.97 & 60.16 & 50.26 & 53.91 & 53.91 \\ 
448K & 33.33 & 50.00 & 48.70 & 60.94 & 59.38 \\ 
896K & 21.62 & 49.22 & 49.74 & 55.47 & 51.56 \\
\bottomrule
\end{tabular}}
\end{table}

\par We performed an additional experiment in which the total memory capacity is fixed at 4096 tokens across all settings. Accordingly, the per-head capacity is 2048, 1024, 512, and 256 tokens for H = 2, 4, 8, and 16, respectively. 

\par The new results (see Table~\ref{tab:mrr-fix-total} and Table~\ref{tab:acc-fix-total}) demonstrate that multi-head partitioning remains highly effective even when the total memory capacity is held constant. Memory retention exhibits a similar trend, increasing from 26.53\% at 896K with H = 1 to 80.68\% (H=2), 73.96\% (H=4), 87.50\% (H=8), and 96.25\% (H=16), respectively. Interestingly, H = 16 achieves the highest memory retention rate (96.25\%), whereas H = 8 achieves the highest end-to-end accuracy (55.47\%). This suggests a trade-off between stronger overwrite protection (more heads) and the representational capacity of each individual head (smaller per-head memory).

\subsection{End-to-end Performance Results on Larger Models}
\begin{table}[htp]
\centering
\caption{Accuracy (\%) results of larger models on RULER-HQA.}\label{tab:acc-larger-models}
\begin{tabular}{llrrr}
\toprule
\multirow{2}{*}{\textbf{Model}} & \multirow{2}{*}{\textbf{Method}} & \multicolumn{3}{c}{Length} \\ \cmidrule{3-5}
 &  & 224K & 448K & 896K \\ \midrule
\multirow{2}{*}{gpt-oss-120b}& LLM Baseline & 43.30 & 28.26 & 22.92 \\ 
 & MHM-LRU & \textbf{58.59} & \textbf{51.56} & \textbf{52.76} \\ \midrule
\multirow{2}{*}{Qwen2.5-32B-Instruct} & LLM Baseline & 25.00 & 13.28 & 7.81 \\ 
 & MHM-LRU & \textbf{60.94} & \textbf{57.81} & \textbf{51.56} \\ \bottomrule
\end{tabular}
\end{table}
\par We added experiments with GPT-OSS-120B and Qwen2.5-32B-Instruct, using their maximum 128K context. The evaluations are based on RULER-HQA. As shown in Table~\ref{tab:acc-larger-models}, both base LLMs degrade sharply at larger context lengths, while MHM-LRU maintains stronger performance.

\subsection{Failure Case Analysis of MHM-LRU}
\label{app:mhm-lru-failure}

\par A failure case analysis of MHM-LRU is shown below. In this case, the agent previously captured the correct answer at step 74 of the memory iteration. However, the agent discarded this correct information at step 78.

\begin{memorycase}{Failure Case: BABILong-512K}
\textbf{Item ID:} 2 \par
\textbf{Question:} Where is Mary? \par
\textbf{Correct answer:} hallway \par
\textbf{Predicted answer:}
Mary is currently traveling towards California with Wayland
Norcross and Berrie McFarlane in Wayland's father's private \ldots \par
\textbf{Number of chunks:} 99 \par
\textbf{Answer present in at least one head:} 6 out of 99 steps
\end{memorycase}

\par The memory capture step is shown below. The bold text highlights the captured answer, while the underline highlights the relevant reasoning chain of the model:

\begin{memorycase}{Step 74: Correct Answer Captured}
\texttt{<memory\_3>}\par
Mary is studying medicine in Glasgow. Previously, she was known to be in the cottage preparing for a party with other characters and later journeyed to the office. Currently, no new location update for Mary has been provided other than her being in Glasgow to study medicine. During the recent section provided, which focuses on narratives about the Inquisition and the experiences of individuals subjected to its cruelty and injustice, \uline{there is a brief mention indicating that Mary moved to the \textbf{hallway}. However, this detail seems out of context with the surrounding narrative and likely pertains to a different individual's experience. Therefore, this movement to the \textbf{hallway} is not considered reliable for locating Mary, who is currently understood to be in Glasgow for her studies.} Source context: Found within biographical narratives and historical testimonials centered around female medical students' experiences in Glasgow around the late 19th century and detailed discussions on the evolution of the British Constitution, intertwined with historical accounts of medieval governance and legal practices, and narratives of the Inquisition's activities.
\par\texttt{</memory\_3>}
\end{memorycase}

\par Although the model captures the "hallway" in its third memory head, the reasoning chain shows that this information is highly uncertain, as it is out of context, surrounded by distraction paragraphs. Therefore, the model abandoned the correct answer in Step 78, causing the failure outcome, as shown below:

\begin{memorycase}{Step 78: Correct Answer Discarded}
\texttt{<memory\_3>}\par
Mary is studying medicine in Glasgow. Previously, she was known to be in the cottage preparing for a party with other characters and later journeyed to the office. Currently, no new location update for Mary has been provided other than her being in Glasgow to study medicine. The recent section provided does not contain any direct information about Mary's current location. This passage includes narratives unrelated to Mary's situation and revolves around personal correspondences and family dynamics, particularly involving a character named Pauline writing letters and her interactions with family members regarding a relative named Hilary. Source context: Found within biographical narratives and historical testimonials centered around female medical students' experiences in Glasgow around the late 19th century and detailed discussions on the evolution of the British Constitution, intertwined with historical accounts of medieval governance and legal practices, and narratives of the Inquisition's activities.
\par\texttt{</memory\_3>}
\end{memorycase}
\section{Memory Head Diversity Analysis}
\subsection{Experiment Setup}
\par An inherent research question for the multi-head recurrent memory agent is: \textbf{\textit{Are memory heads differentiated or not?}} To answer this question, we propose a semantic distance-based analysis of memory head diversity, demonstrating whether and to what extent memory heads semantically differ from one another.

\paragraph{Evaluation Metric.} We measure the diversity of memory heads using the average cosine distance for memory states at each time step. Specifically, given a memory trajectory of a test item, denote $m_t = \{m_t^1,\ldots,m_t^H\}$ as the memory state of MHM at time step $t$. We first use a sentence transformer (\texttt{all-mpnet-base-v2} in the experiment) to obtain the semantic vector of each memory head, i.e.,
\begin{equation}
    v_t^h=\text{SentenceTransformer}(m_t^h),h\in[H].
\end{equation}
\par Next, we calculate the average memory vector $\overline{v}_t = \frac{1}{H}\sum_{h=1}^H v_t^h$. We define the cosine distance between each memory head and the averaged memory by:
\begin{equation}
    d_{\cos}(v_t^h, \overline{v}_t) = 1-\cos(v_t^h, \overline{v}_t),
\end{equation}
\par Finally, the \textbf{averaged cosine distance $\overline{d}_t$} is defined by:
\begin{equation}
    \overline{d}_t = \frac{1}{H}\sum_{h=1}^H  d_{\cos}(v_t^h, \overline{v}_t)
\end{equation}
\par The $\overline{d}_t$ is non-negative and measures the diversity of memory heads at time step $t$. When $\overline{d}_t=0$, all memory heads are semantically identical. The higher the $\overline{d}_t$, the more diverse the memory heads are. For each memory trajectory, we visualize $\overline{d}_t$ with respect to time step $t$ to see how memory diversity evolves as time goes by.

\subsection{Experiment Results}
\par
\begin{figure}[htp]
    \centering
    \begin{subfigure}{\linewidth}
        \includegraphics[width=\linewidth]{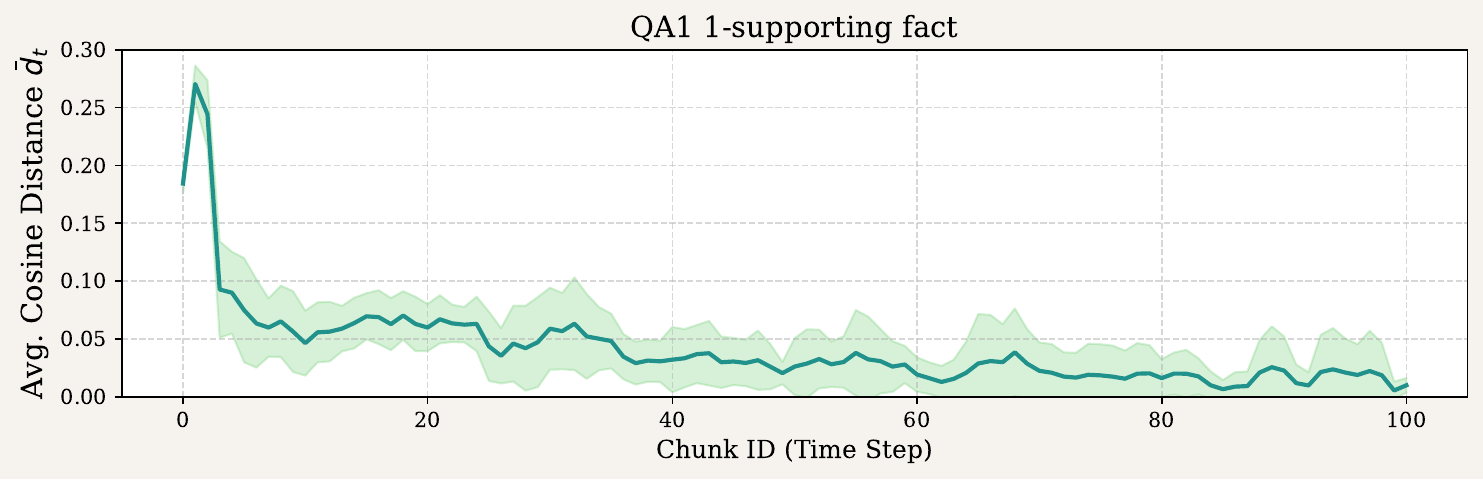}
        \caption{QA1: 1-supporting fact.}
    \label{fig:memory-diversity-qa1}
    \end{subfigure}
    
    \begin{subfigure}{\linewidth}
        \includegraphics[width=\linewidth]{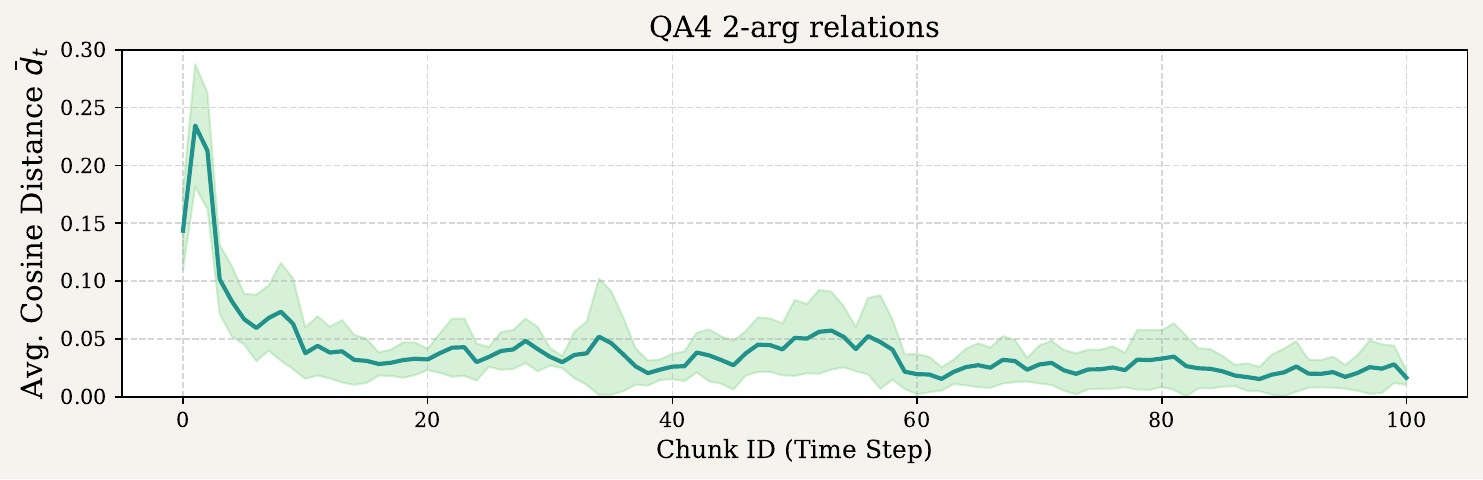}
        \caption{QA4: 2-arg relations.}
    \label{fig:memory-diversity-qa4}
    \end{subfigure}

    \begin{subfigure}{\linewidth}
        \includegraphics[width=\linewidth]{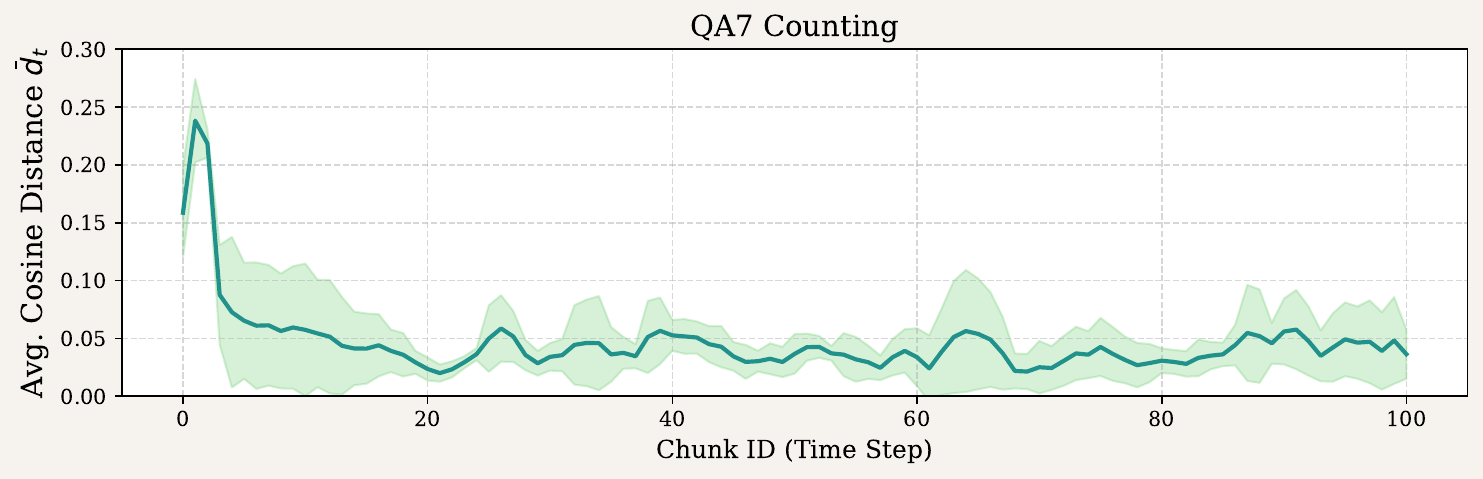}
        \caption{QA7: Counting.}
    \label{fig:memory-diversity-qa7}
    \end{subfigure}
    \caption{Average memory head distance ($\overline{d}_t$) with respect to time steps. The higher the $\overline{d}_t$, the more semantically diverse the memory heads in the time step.}
    \label{fig:memory-diversity}
\end{figure}

 We evaluate memory diversity on three representative tasks in BABILong: the QA1 (1-supporting fact), QA4 (2-arg relations) and QA7 (Counting), choosing five trajectories with continuous item IDs for visualization for each task. Results are shown in Figure~\ref{fig:memory-diversity}.
\paragraph{MHM Enables Diverse Memory Allocation.} Figure~\ref{fig:memory-diversity} shows that MHM-LRU consistently produces non-trivial semantic diversity across memory heads throughout processing. Across all three tasks, $\overline{d}_t$ is positive at most time steps, confirming that heads spontaneously specialize without any explicit diversity objective. Diversity is highest in the initial steps when each head is first written, then gradually decreases and stabilizes at a lower plateau as MHM-LRU propagates task-relevant information redundantly across heads—the same emergent backup behavior documented in Section~\ref{sec:case-study}. This trajectory reflects a natural balance between early-stage coverage and late-stage retention robustness.

\paragraph{Diversity Varies In Different Tasks.} The magnitude and duration of head diversity are modulated by task complexity. For QA1 (1-supporting fact), $\overline{d}_t$ collapses rapidly once the single target fact is captured, with heads converging to near-identical content by step 20–40. QA4 (2-arg relations) sustains higher diversity for longer. QA7 (Counting) exhibits the most persistent and fluctuating diversity, consistent with the incremental nature of counting where each new chunk may modify the running tally. This pattern aligns with the retention gains in Section~\ref{sec:utility}: tasks requiring multi-step information accumulation benefit most from heads that remain semantically differentiated over longer processing horizons.

\section{Broader Impact}~\label{app:impact}
\par In this section, we discuss the broader impact of MHM not only on recurrent memory for long-context query, but more general research field of agentic memory in the future.

\paragraph{Towards Long-term and Heterogeneous Memory.} Despite the success of MHM in single-turn long-context query, the current research focus of recurrent memory is still limited by query-specific, short-term memory. An inevitable following research question is: what a long-term memory should be, and how to maintain it. Long-term memory is valuable in the whole life cycle of an agent, containing a variety of heterogeneous memory information, and is query-agnostic. The multi-head memory has the potential to achieve this: by allowing each memory head to specialize in storing a distinct type of information, such as episodic experience, declarative knowledge, procedural skills, or relational context, MHM provides a natural architectural substrate for heterogeneous long-term memory. Unlike the monolithic single-head design, which is forced to compress all memory types into a single undifferentiated representation, the multi-head structure admits a principled partition of memory responsibilities across heads. This partitioning can be further guided by semantic-aware or type-aware selection strategies, enabling the agent to organize, retrieve, and update memory content in a structured and persistent manner across its entire operational lifetime.

\paragraph{Towards Self-Evolving Agent.} We believe that a must-go path for memory agent is to achieve a realistic self-evolving agent---the agent can learn experience from past failure or success, and accumulate knowledge from observation and simulation. The multi-head memory essentially provides a technical path towards the self-evolving agent: by endowing agents with the ability to spontaneously and freely manage and utilize multiple memory heads, the agent can achieve self-evolvement of its experience through self-evolvement of memory internalization. In the long run, we envision that MHM-style architectures, when coupled with continual learning and reflection mechanisms, could enable agents to autonomously consolidate short-term observations into long-term behavioral priors, progressively refine their internal representations in response to environmental feedback, and ultimately exhibit sustained improvement over time without human intervention, which is a foundational step toward truly autonomous and adaptive AI systems.

\newpage
\section*{NeurIPS Paper Checklist}

\begin{enumerate}

\item {\bf Claims}
    \item[] Question: Do the main claims made in the abstract and introduction accurately reflect the paper's contributions and scope?
    \item[] Answer: \answerYes{} % Replace by \answerYes{}, \answerNo{}, or \answerNA{}.
    \item[] Justification: The main claims made in the abstract and introduction accurately reflect the paper's contribution on recurrent memory agent, long context LLMs and their broader range of application.

\item {\bf Limitations}
    \item[] Question: Does the paper discuss the limitations of the work performed by the authors?
    \item[] Answer: \answerYes{} % Replace by \answerYes{}, \answerNo{}, or \answerNA{}.
    \item[] Justification: The limitations of the work are discussed in Section~\ref{sec:conclusion}

\item {\bf Theory assumptions and proofs}
    \item[] Question: For each theoretical result, does the paper provide the full set of assumptions and a complete (and correct) proof?
    \item[] Answer: \answerYes{} % Replace by \answerYes{}, \answerNo{}, or \answerNA{}.
    \item[] Justification: Full set of assumptions, correct and complete proof, are presented in the appendix.

    \item {\bf Experimental result reproducibility}
    \item[] Question: Does the paper fully disclose all the information needed to reproduce the main experimental results of the paper to the extent that it affects the main claims and/or conclusions of the paper (regardless of whether the code and data are provided or not)?
    \item[] Answer: \answerYes{} % Replace by \answerYes{}, \answerNo{}, or \answerNA{}.
    \item[] Justification: Experiment settings are demonstrated in Section~\ref{sec:experiment} and Appendix~\ref{app:experiment-setting}.

\item {\bf Open access to data and code}
    \item[] Question: Does the paper provide open access to the data and code, with sufficient instructions to faithfully reproduce the main experimental results, as described in supplemental material?
    \item[] Answer: \answerYes{} % Replace by \answerYes{}, \answerNo{}, or \answerNA{}.
    \item[] Justification: Experiment data is publicly available online. Code will be released publicly upon publication.

\item {\bf Experimental setting/details}
    \item[] Question: Does the paper specify all the training and test details (e.g., data splits, hyperparameters, how they were chosen, type of optimizer) necessary to understand the results?
    \item[] Answer: \answerYes{} % Replace by \answerYes{}, \answerNo{}, or \answerNA{}.
    \item[] Justification: Experiment settings are presented in Appendix~\ref{app:experiment-setting}.

\item {\bf Experiment statistical significance}
    \item[] Question: Does the paper report error bars suitably and correctly defined or other appropriate information about the statistical significance of the experiments?
    \item[] Answer: \answerYes{} % Replace by \answerYes{}, \answerNo{}, or \answerNA{}.
    \item[] Justification: We report the mean and standard deviation for evaluation results in main experiments.

\item {\bf Experiments compute resources}
    \item[] Question: For each experiment, does the paper provide sufficient information on the computer resources (type of compute workers, memory, time of execution) needed to reproduce the experiments?
    \item[] Answer: \answerYes{} % Replace by \answerYes{}, \answerNo{}, or \answerNA{}.
    \item[] Justification: We report the type of computation workers, required memory and time of execution in the utility analysis.
    
\item {\bf Code of ethics}
    \item[] Question: Does the research conducted in the paper conform, in every respect, with the NeurIPS Code of Ethics \url{https://neurips.cc/public/EthicsGuidelines}?
    \item[] Answer: \answerYes{} % Replace by \answerYes{}, \answerNo{}, or \answerNA{}.
    \item[] Justification: The research confirm with the NeurIPS Code of Ethics.

\item {\bf Broader impacts}
    \item[] Question: Does the paper discuss both potential positive societal impacts and negative societal impacts of the work performed?
    \item[] Answer: \answerYes{} % Replace by \answerYes{}, \answerNo{}, or \answerNA{}.
    \item[] Justification: The paper discusses the potential social impacts in Appendix~\ref{app:impact}.
    
\item {\bf Safeguards}
    \item[] Question: Does the paper describe safeguards that have been put in place for responsible release of data or models that have a high risk for misuse (e.g., pre-trained language models, image generators, or scraped datasets)?
    \item[] Answer: \answerNA{} % Replace by \answerYes{}, \answerNo{}, or \answerNA{}.
    \item[] Justification: Data and models used in this paper are fully open-sourced and available on the Internet. %\justificationTODO{}

\item {\bf Licenses for existing assets}
    \item[] Question: Are the creators or original owners of assets (e.g., code, data, models), used in the paper, properly credited and are the license and terms of use explicitly mentioned and properly respected?
    \item[] Answer: \answerYes{} % Replace by \answerYes{}, \answerNo{}, or \answerNA{}.
    \item[] Justification: All the data, code and model used in this paper are open-sourced and we rigorously obey the license and terms of use.

\item {\bf New assets}
    \item[] Question: Are new assets introduced in the paper well documented and is the documentation provided alongside the assets?
    \item[] Answer: \answerYes{} % Replace by \answerYes{}, \answerNo{}, or \answerNA{}.
    \item[] Justification: The new assets (mainly code) are well documented.

\item {\bf Crowdsourcing and research with human subjects}
    \item[] Question: For crowdsourcing experiments and research with human subjects, does the paper include the full text of instructions given to participants and screenshots, if applicable, as well as details about compensation (if any)? 
    \item[] Answer: \answerNA{} % Replace by \answerYes{}, \answerNo{}, or \answerNA{}.
    \item[] Justification: This research does not involve crowdsourcing nor research with human subjects.

\item {\bf Institutional review board (IRB) approvals or equivalent for research with human subjects}
    \item[] Question: Does the paper describe potential risks incurred by study participants, whether such risks were disclosed to the subjects, and whether Institutional Review Board (IRB) approvals (or an equivalent approval/review based on the requirements of your country or institution) were obtained?
    \item[] Answer: \answerNA{} % Replace by \answerYes{}, \answerNo{}, or \answerNA{}.
    \item[] Justification: This research does not involve crowdsourcing nor reearch with human subjects.

\item {\bf Declaration of LLM usage}
    \item[] Question: Does the paper describe the usage of LLMs if it is an important, original, or non-standard component of the core methods in this research? Note that if the LLM is used only for writing, editing, or formatting purposes and does \emph{not} impact the core methodology, scientific rigor, or originality of the research, declaration is not required.
    \item[] Answer: \answerNA{} % Replace by \answerYes{}, \answerNo{}, or \answerNA{}.
    \item[] Justification: The core method development in this research does not involve LLMs as any important, original or non-standard components.

\end{enumerate}

\end{document}